\documentclass[10pt, a4paper]{article}

\usepackage{lrec-coling2024} % this is the new style

\usepackage{times}
\usepackage{latexsym}
\usepackage{graphicx}
\usepackage{algorithm}
\usepackage{algorithmic}
\usepackage{amsmath}
\usepackage{multirow}
\usepackage{makecell}
\usepackage{amssymb}
\usepackage{booktabs}
\usepackage{pifont}

\title{TacoERE: Cluster-aware Compression for Event Relation Extraction}

\name{Yong Guan$^{1}$, Xiaozhi Wang$^{1}$, Lei Hou$^{1}$$^{*}$\thanks{*Corresponding author}, Juanzi Li$^{1}$, Jeff Pan$^{2}$, \\ {\bf \large Jiaoyan Chen$^{3}$, Freddy Lecue$^{4}$}}

\address{$^1$Department of Computer Science and Technology, Tsinghua University, Beijing, China \\
         $^2$School of Informatics, The University of Edinburgh, UK \\
         $^3$Department of Computer Science, The University of Manchester, UK \\ 
         $^4$INRIA, France \\ 
         \{gy2022, wangxz20\}@mails.tsinghua.edu.cn,\ \  \{houlei, lijuanzi\}@tsinghua.edu.cn\\}

\abstract{
Event relation extraction (ERE) is a critical and fundamental challenge for natural language processing. Existing work mainly focuses on directly modeling the entire document, which cannot effectively handle long-range dependencies and information redundancy. To address these issues, we propose a clus\underline{T}er-\underline{a}ware \underline{co}mpression method for improving \underline{E}vent \underline{R}elation \underline{E}xtraction (TacoERE), which explores a compression-then-extraction paradigm. Specifically, we first introduce document clustering for modeling event dependencies. It splits the document into intra- and inter-clusters, where intra-clusters aim to enhance the relations within the same cluster, while inter-clusters attempt to model the related events at arbitrary distances. Secondly, we utilize cluster summarization to simplify and highlight important text content of clusters for mitigating information redundancy and event distance. We have conducted extensive experiments on both pre-trained language models, such as RoBERTa, and large language models, such as ChatGPT and GPT-4, on three ERE datasets, i.e., MAVEN-ERE, EventStoryLine  and HiEve. Experimental results demonstrate that TacoERE is an effective method for ERE.
 \\ \newline \Keywords{Event Relation Extraction, Compression-then-Extraction, Large Language Model} }

\begin{document}

\maketitleabstract

\section{Introduction}
\label{sec_introduction}

\begin{figure}[t]
  \centering
  \includegraphics[width=\linewidth]{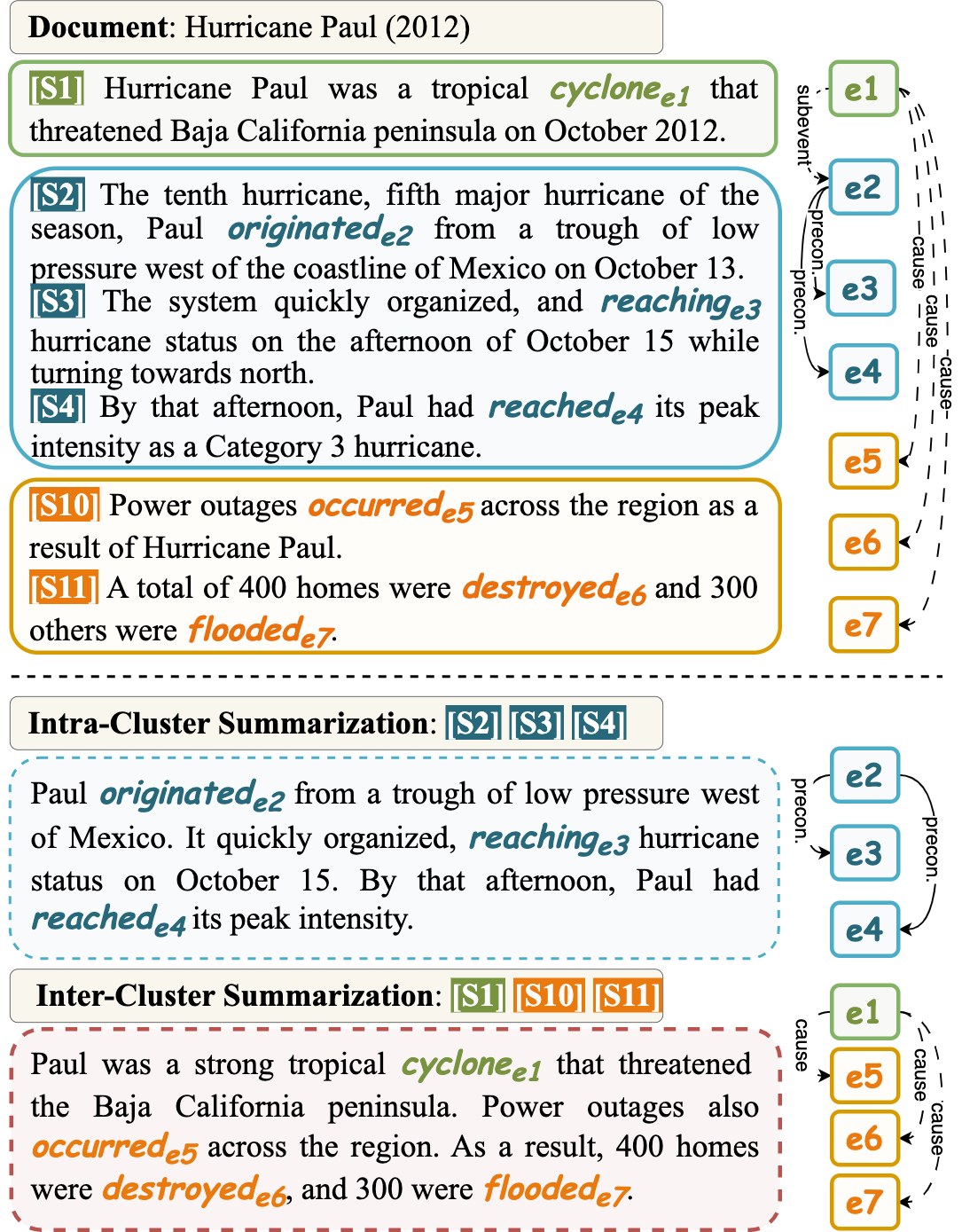}\\
  \caption{An example form MAVEN-ERE. Words in bold italics are trigger words of events. [S\textit{i}] denotes the \textit{i}-th sentence index, and precon. is the abbreviation of precondition. The solid and dashed arrows indicate relations within and among clusters, respectively. Different colored sentences represent different clusters.}
  \label{intor_example}
\end{figure}

Event Relation Extraction (ERE) aims to predict relations, such as causal and subevent relations, between event mentions or trigger words in a document \cite{fan_eci_2022}. 
As shown in Figure \ref{intor_example}, given a document with event mentions/trigger words, an ERE model is expected to predict relations among the three mentioned events, such as $\texttt{cyclone} (\textbf{e1})  \xrightarrow[]{subevent} \texttt{originated} (\textbf{e2}) \xrightarrow[]{precondition} \texttt{reached} (\textbf{e4})$. 
ERE can not only facilitate deep understanding of text \cite{wang-etal-2020-joint}, but also benefit various downstream tasks, such as question answering \cite{khashabi-2018-qa} and information retrieval \cite{pang2020setrank}.

With the widespread adoption of deep neural networks in natural language processing (NLP), event relation extraction systems have undergone a paradigm shift to supervised neural models that encode the document as a clue for predicting relations~\cite{cao-etal-2021-knowledge,xu-etal-2021-discriminative,chen-etal-2022-ergo}. 
However, there are still two challenges \textbf{long-range dependencies} and \textbf{information redundancy}. 
Specifically, \textit{long-range dependencies} indicates that events may be scattered across multiple sentences potentially far away from each other. In such context, existing ERE methods have difficulties capturing the dependencies among events. 
Consider example in Figure \ref{intor_example}, events \texttt{cyclone} (\textbf{e1}) and \texttt{destroyed} (\textbf{e7}), located respectively in sentence S1 and S11, are related with a \texttt{cause} relation. 
\textit{Information redundancy} refers to the existence of information non relevant for relation prediction. For example, identifying the relation between events \texttt{originated} (\textbf{e2}) and \texttt{reaching} (\textbf{e3}) only depends on the sentences S2 and S3, while sentences S9, S10, and S11 are non relevant for identifying the relation.

To tackle these challenges, the major approaches currently are to select sentences~\cite{wang-etal-2020-joint,Man_Ngo_Van_Nguyen_2022} or remove sentences~\cite{xu-etal-2022-document}. But these methods do not completely eliminate long range dependencies, or presence of irrelevant information at finer granularities (e.g., clauses/phrases). As shown in Figure \ref{intor_example}, for long-range dependencies, 
deletes sentences S6, S7, and S8, but the distance between sentences S1 and S11 remains considerable, indicating that the dependency on sentence distance remains present.
For information redundancy, the sub-sentence of sentence S2, "\textit{tenth hurricane, and fifth major hurricane of the season}", is still useless for predicting the relation between \texttt{originated} (\textbf{e2}) and \texttt{reaching} (\textbf{e3}) from human understanding. This motivates our hypothesis that compression via summarization might be a better strategy than sentence filtering.

In this paper, we propose TacoERE, a clus\underline{T}er-\underline{a}ware \underline{co}mpression method for improving \underline{E}vent \underline{R}elation \underline{E}xtraction, which explores a compression-then-extraction paradigm to extract event relations. 
Specifically, TacoERE first uses document clustering to split the document into intra- and inter-clusters\footnote{Document generally utilizes multiple sub-topics to organise the content~\cite{topic_hearst_1993}. As shown in Figure \ref{intor_example}, the three sub-topics, event background, development process, and resulting impact, collectively constitute the entire article. For simplified understanding, we use the term ``cluster'' to represent sentences within the same sub-topic.}, where intra-clusters aim to enhance the relations within the same cluster, while inter-clusters attempt to model the related events at arbitrary distances.
For instance, events \texttt{cyclone} (\textbf{e1}) in sentence S1 and \texttt{destroyed} (\textbf{e7}) in sentence S11 belong to different clusters, and combining these two clusters can effectively reduce the event distance and facilitate modelling event relations.
In this way, all dependencies between related events of any distance in the document can be modelled. 
Following that, cluster summarization is adopted to generate summaries for every cluster, which can further simplify and highlight important content. 
At last, the generated summaries of intra- and inter-clusters are utilized to predict relations.

For evaluation, we respectively validate our ideas on both small-scale pre-trained language models (PLMs), such as RoBERTa, and large language models (LLMs), such as ChatGPT\footnote{ \url{https://chat.openai.com/chat}} and GPT-4 \cite{openai2023gpt4}, and conduct extensive experiments on three ERE datasets, namely, MAVEN-ERE~\citeplanguageresource{MAVEN-ERE}, EventStoryLine~\citeplanguageresource{EventStoryLine} and HiEve~\citeplanguageresource{HiEve}. Experimental results demonstrate that our approach can effectively improve the performance of event relation extraction models. Our contributions are summarized as follows:

\begin{itemize}
    \item We propose a novel cluster-aware compression method for event relation extraction, namely, TacoERE, which explores a compression-then-extraction paradigm to extract relations.
    \item We utilize document clustering to split the document into intra- and inter-clusters to allow the modeling of dependencies without any reliance on event distance. We propose cluster summarization to simplify and spotlight important text content to mitigate the impact of information redundancy and event distance.
    \item Extensive experiments have been conducted on both PLMs, such as RoBERTa, and LLMs, such as ChatGPT and GPT-4, on three ERE datasets, i.e., MAVEN-ERE, EventStoryLine and HiEve. Our TacoERE outperforms existing methods, especially on LLMs, with improvements by 11.2\% and 9.1\% on ChatGPT and GPT-4 respectively.
\end{itemize}

\begin{figure*}[!htp]
  \centering
  \includegraphics[width=0.95\linewidth]{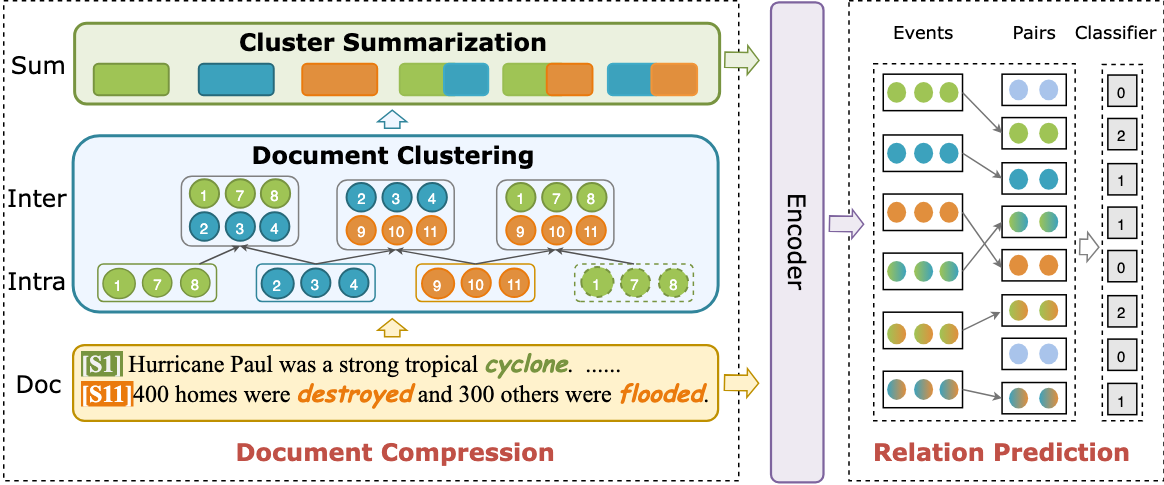}\\
  \caption{Model structure of TacoERE.}
  \label{meth_struc}
\end{figure*}

\section{TacoERE}
\label{sec_methodology}

We consider a document   $D = \{ s_{1}, s_{2}, ..., s_{n}\}$ with $n$ sentences, annotated with event mentions/trigger words $E = \{e_{1}, e_{2}, ..., e_{m}\}$. 
The task of event relation extraction is to, given an annotated document $D$ and a pre-defined relation set $\mathcal{R}$, predict relations between all event pairs $\{(e_{i}, e_{j})\}$  in $D$.

In this paper, we propose TacoERE, a novel cluster-aware compression method for improving event relation extraction, exploring a compression-then-extraction paradigm to extract relations. Figure \ref{meth_struc} shows the overview of our framework: we first present \textit{Document Compression}, including \textit{Document Clustering} (Section \ref{sec_topic_grouping}) and \textit{Cluster Summarization} (Section \ref{sec_topic_compression}). 
\textit{Document Clustering} splits document into intra- and inter-clusters to allow the modeling of dependencies without considering event distance among sentences. \textit{Cluster Summarization} encourages model to simplify and spotlight important content of clusters for mitigating information redundancy and event distance. 
We then describe \textit{Relation Prediction} in Section \ref{sec_relationprediction}, which utilizes the content from \textit{Cluster Summarization} to predict relations. 
Last, in Section \ref{sec_training}, we employ a reinforce algorithm that jointly optimizes the \textit{Cluster Summarization} and \textit{Relation Prediction}. 
Before joint training, we introduce pretraining module for \textit{Cluster Summarization} with event chains to teach model to do better content representation.

\subsection{Document Clustering}
\label{sec_topic_grouping}

The document clustering aims to split document into intra- and inter-clusters, where intra-clusters aim to enhance the relations within the same cluster, while inter-clusters attempt to model the related events at arbitrary distances. 
Our observation is that predicting the relations for event pairs relies on a limited content rather than the entire document.
For example in Figure \ref{intor_example}, just based on sentence S2 and S3, the relation between event \texttt{originated} (\textbf{e2}) and \texttt{reaching} (\textbf{e3}) can be deduced.
Moreover, sentences in document with the same cluster are more probably to involve event relations, such as the cluster of development process with blue color (sentences S2-S4) in Figure \ref{intor_example} has three relations.
As a result, we split the $D$ into $K$ mutually independent intra-clusters at sentence level. To obtain the intra-clusters, arbitrary clustering methods, such as traditional machine learning or deep neural networks, can be applied. In our experiments, we directly utilize the effective and widely used K-means algorithm \cite{cluster_guan_2020,rakib_cluster_2020}. 
Specifically, we extract multiple features to enhance the clustering, including cluster words extracted by LDA model, trigger words, tf-idf, and sentence representation encoded by RoBERTa.
\begin{equation}
    J = \sum^{k}_{i=1} \sum_{j=1}p_{ij}|| v_{j} - u_{i} ||^{2}
\end{equation}
where $J$ is objective, $v_{j}$ is the feature of sentence $s_{j}$, $u_{i}$ is center of \textit{i}-th cluster, and $p_{ij}$ is indicator.

Beyond individual intra-cluster, among different intra-clusters also occur with event relations which can be seen in Figure \ref{intor_example}, relation between event \texttt{cyclone} (\textbf{e1}) in background and \texttt{destroyed} (e7) in impact. Thus, we fuse any two intra-clusters as inter-clusters. 
The order of sentences in intra- and inter-clusters will follow their original order in $D$.
And the intra- and inter-clusters can involve all possible relations in $D$.

\subsection{Cluster Summarization}
\label{sec_topic_compression}

The cluster summarization aims to simplify and highlight important text content of clusters for mitigating information redundancy and event distance. 
Intra- and inter-clusters directly group the sentences with related sub-topics. It will inevitably contain redundant information and ignore coherence, which may hinder the performance \cite{gao-etal-2021-making}.
Text summarization as a technique can effectively simplify and spotlight important text content, which would solve this problem to some extent. 
To this end, we further utilize a summarization model to generate summaries $C^{a}$ and $C^{r}$ for intra- and inter-clusters respectively, where $\{ C^{a}, C^{r} \} \in C$. In this paper, we utilize a transformer-based \cite{vaswani2017attention} encoder-decoder framework for summarization. In particular, a pretrained language model as the encoder to learn the contextual representation of input, and a transformer-based decoder is utilized to generate its summary word-by-word.

\subsection{Relation Prediction}
\label{sec_relationprediction}

The relation prediction aims to construct the event relation prediction process by giving the text content from cluster summarization. We first need to obtain the contextual representation of each token in the document $D$, $C^{a}$, and $C^{r}$ respectively. Take $D$ as an example, we leverage pre-trained language model RoBERTa \cite{liu2019roberta} as the encoder.
Since event mention/trigger words perhaps contain  multiple words, i.e., \textit{take place}, and individual word in document may be split into sub-words by wordpiece, i.e., word ``\textit{summarization}'' will be split into three sub-words ``\textit{sum}'', ``\textit{mar}'', and ``\textit{ization}'', we adopt LogSumExp pooling method \cite{Zhou_Huang_Ma_Huang_2021} over all its mentions (sub-words) embeddings in the last encoding layer as the event representation. Then, we can obtain the overall representation of event pairs in $D$.
For cluster summaries $C^{a}$ and $C^{r}$, we extract the events which appear in original document $D$. 

Event pairs of $C^{a}$ and $C^{r}$ all appear in $D$, which means one event pair may occur multiple times. Different from existing work \cite{adhikari2019docbert,beltagy2020longformer} which aggregate all identical event representations to get the final event pair representations, for the same event pair, we select the most relevant one, and the detailed selection process is as follows: (1) for the same event pair, the priority of event pair in $C^{r}$ is higher than $C^{a}$; (2) for the event pair which not in $C^{a}$ and $C^{r}$, we directly use the representation in $D$. Finally, a two layers feed-forward network with softmax is adopted to learn the classes probability based on the event pair representations. For training objective of relation prediction $\mathcal{L}_{rp}$, we use cross-entropy function as follows:
\begin{equation}
    \mathcal{L}_{rp} = - \sum_{i \neq j} \sum_{\mathcal{R}} \{ r_{ij} \mathit{log} P_{ij} + (1-r_{ij}) \mathit{log}(1-P_{ij}) \}
\end{equation}
where $P_{ij}$ is the probabilities between events $e_{i}$ and $e_{j}$, and $r_{ij} \in \mathcal{R}$.

\subsection{Training Phase}
\label{sec_training}

The Training Phase describes both \textit{Joint Training} and \textit{Pretraining for Cluster Summarization}, where \textit{Joint Training} aims to jointly optimize the cluster summarization and relation prediction, while \textit{Pretraining for Cluster Summarization} aims to teach model to do better content representation. 

\noindent \textbf{Joint Training} \enspace
As the performance cannot be directly back-propagated to the cluster summarization process, we employ a reinforce algorithm \cite{williams-1992} that treats the event relation prediction performance as reward function for the cluster summaries to train cluster summarization process. Besides, we also consider another event chains information from clusters and its summaries to enrich the overall reward function $\mathcal{R}(C)$.

\textit{Performance-based Reward} $\mathcal{R}^{per}(C)$ aims to use the performance as the direct training signal. We calculate the reward $\mathcal{R}^{per}(C)$ based on the performance for all document event pairs.
In particular, for each event pair $(e_{i}, e_{j})$, $\mathcal{R}^{per}(C)$ = 1 if relation prediction model calculates the true relations between $e_{i}$ and $e_{j}$, and 0 otherwise.

\textit{Summary-based Reward} $\mathcal{R}^{ec}(C)$ aims to encourage the correlation between the clusters and its summaries to train cluster summarization.
The motivation of $\mathcal{R}^{ec}(C)$ is that summary expresses important information about the text and they are consistent at the semantic level \cite{GUAN2021106973}. 
In Section \ref{sec_topic_grouping}, we split document into intra- and inter-clusters. Intra-clusters are independent of each other and cover the content of the entire document. Thus, we only adopt the intra-clusters and its summaries to calculate the reward $\mathcal{R}^{ec}(C)$. Similar to existing work~\citet{pasunuru-bansal-2018-multi,paulus2018a}, we utilize the popularly known automatic evaluation metric for summarization, ROUGE~\cite{lin-2004-rouge}, as reward function. However, this metric mainly focuses on phrase matching/n-gram overlap while assuming equal contributions from each word. Addressing these issues, we introduce salient function to give higher weight to the trigger words. 
\begin{equation}
    P = \frac{\sum_{k} ( \mathit{LCS_{\cup}} (C^{a}_{k}, T^{a}_{k})+ \sum_{l} \eta(w_{kl}))}{|D|}
\end{equation}
\begin{equation}
    R = \frac{\sum_{k} ( \mathit{LCS_{\cup}} (C^{a}_{k}, T^{a}_{k})+ \sum_{l} \eta(w_{kl}))}{\sum_{k}|C^{a}_{k}|}
\end{equation}
\begin{equation}
    \mathcal{R}^{ec}(C) = \frac{(1+\sigma^{2})RP}{R+\sigma^{2}P}
\end{equation}
where $C^{a}_{k}$ is the summary of $k$-th intra-cluster $T^{a}_{k}$, $\mathit{LCS_{\cup}(\cdot)}$ is the union longest common subsequence, $\sigma$ is defined in~\citet{lin-2004-rouge}, and $\eta(\cdot)$ is function to measure whether $w_{kl}$ is a trigger word. 

After obtaining reward $\mathcal{R}^{per}(C)$ and $\mathcal{R}^{ec}(C)$, the overall reward can be calculated as $\mathcal{R}(C) = \alpha  \mathcal{R}^{per}(C) + \beta \mathcal{R}^{ec}(C)$, where $\alpha$ and $\beta$ are trade-off parameters.
 Following existing work \cite{Man_Ngo_Van_Nguyen_2022}, we minimize the negative expected reward $\mathcal{R}(C)$ over the possible choices of summaries:
\begin{equation}
    \mathcal{L}_{sum} = - \mathbb{E}_{C^{\prime} \sim P(C^{\prime} | e_{i}, e_{j}, D)} [\mathcal{R}(C^{\prime})]
\end{equation}
Then, the gradient can be further formalised by utilizing the one roll-out sample method.
\begin{equation}
    \nabla \mathcal{L}_{sum} = -(\mathcal{R}(C) - \theta)\nabla \mathit{log} P(C| e_{i}, e_{j}, D)
\end{equation}
where $\theta$ is used to reduce variance.

\noindent \textbf{Pretraining for Cluster Summarization} \enspace This module aims to teach model to do better content representation for cluster summarization process with event chains before joint training phase.
As presented in Section \ref{sec_topic_compression}, we utilize summarization model to compress the different clusters for predicting the event relations. However, abstractive summarization method always generates new words. How to ensure generated summaries contain the events in the original document is a key problem.
Inspired by \cite{narayan-etal-2021-planning,narayan-etal-2022-well}, we utilize the event chains which order the events in the summary as an intermediate summary representation to better guide the summarization generation.
In particular, we concatenate the event chain with the corresponding summary as a unified sequence, such as ``\textit{[EVENTCHAIN] originated | organized | reaching | reached | ... [SUMMARY] paul originated from a trough of low pressure ...}''.
During the decoder phase, the model must generate both the event chain followed by the summary. 
The model structure is similar to the summarization model in Section \ref{sec_topic_compression}, and we select the existing abstractive summarization datasets, such as CNN/DailyMail, for the pretraining.

 \section{Experiments}

In this section, we first introduce experiment setup (Section \ref{exp_setup}), and then report the results and analysis (Section \ref{exp_overallresults} - \ref{exp_ablationstudy}). 
Furthermore, we also conduct experiments on LLMs in Section \ref{exp_llms}.

\subsection{Experiment Setup}
\label{exp_setup}

\textbf{Datasets} \enspace
We conduct experiments on three datasets: MAVEN-ERE~\citeplanguageresource{MAVEN-ERE}, EventStoryLine~\citeplanguageresource{EventStoryLine} and HiEve~\citeplanguageresource{HiEve}. 
MAVEN-ERE is a unified large-scale human-annotated ERE dataset, which contains 4,480 documents, 112,276 events, 57,992 causal relations, and 15,841 subevent relations in total. We split the data into train, dev, and test sets in 2,913, 710, and 857 documents as in \citet{wang-etal-2022-maven}.
EventStoryLine (i.e., version 0.9) contains 258 documents, 22 topics, 5,334 events, and 5,655 causal relations. Following existing works \cite{gao-etal-2019-modeling,tran-phu-nguyen-2021-graph}, we use the last two topics as development set, and the other 20 topics are verified with a 5-fold cross-validation. For HiEve, the dataset contains 100 documents, 3,185 events, and 3,648 subevent relations. Similar to~\citet{zhou-etal-2020-temporal,wang-etal-2020-joint}, we split the 100 documents into 60 training, 20 validation, and 20 testing.

\noindent \textbf{Evaluation Details} \enspace
We adopt similar settings to~\citet{wang-etal-2022-maven}, which are closer to real situations, yet more challenging, from three aspects:
(1) we consider the relation directions for relation prediction. In addition, causal/subevent relations may be defined in several sub-relations. For example, the causal relation in MAVEN-ERE has two sub-relations ``\textit{CAUSE}'' and ``\textit{PRECONDITION}''; 
(2) we report the overall score rather than the individual sub-relation score; 
(3) we do not down-sample the negative instances.

Following the existing works \cite{gao-etal-2019-modeling,zhou-etal-2020-temporal,tran-phu-nguyen-2021-graph}, we use standard Precision (P), Recall (R), and F1-score (F1) as the metrics.

\begin{table}[t]
    \centering
    \begin{tabular}{l p{0.9cm}<{\centering}p{0.9cm}<{\centering}p{0.9cm}<{\centering}}
         \toprule
        Methods&P&R&F1\\
         \midrule
         \multicolumn{4}{c}{MAVEN-ERE}\\
         \midrule
         BERT&$31.6$&$28.2$&$29.9$\\
         RoBERTa&$33.8$&$29.5$&$31.5$\\
         Hierarchical&$31.8$&$29.2$&$30.6$\\
         SIEF&$33.6$&$30.8$&$32.3$\\
         \midrule
         TacoERE (PLMs)&$\textbf{34.8}$&$\textbf{32.4}$&$\textbf{34.1}$\\
         \midrule
        \multicolumn{4}{c}{EventStoryLine}\\
         \midrule
         BERT&$30.3$&$9.4$&$12.8$\\
         RoBERTa&$31.1$&$10.7$&$14.4$\\
         Hierarchical&$30.1$&$10.2$&$13.1$\\
         SIEF&$32.4$&$11.3$&$14.8$\\
         SCS-EERE&$32.7$&$10.9$&$15.1$\\
         \midrule
         TacoERE (PLMs)&$\textbf{32.9}$&$\textbf{12.3}$&$\textbf{16.4}$\\
         \bottomrule         
    \end{tabular}
    \caption{Model performance of causal relation on MAVEN-ERE and EventStoryLine.}
    \label{tab:causal_per}
\end{table}

\noindent \textbf{Training Details} \enspace
For cluster summarization, we utilize BERT as the document encoder, and a transformer-based framework as the decoder with 2 layers and 8 attention heads.
For document clustering, the value $K$ of intra-cluster is a hyper parameter and we set $K=3$ with the best performance for the experiment. 
For relation prediction, we adopt RoBERTa as the document encoder.
The implementation of BERT and RoBERTa are based on the pytorch version from HuggingFace Transformers library \footnote{\url{https://github.com/huggingface}}.
We adopt adam optimizer with learning rate of 5e-4. The trade-off parameters of $\alpha$ and $\beta$ are set to 1.0 and 0.1 respectively.
Each training and testing process is running on two NVIDIA GeForce RTX 3090 GPU. 

For LLMs implementation, we use the model API provided by OpenAI, where ``gpt-4'', ``gpt-3.5-turbo'', and ``text-davinci-003'' refer to models GPT-4, ChatGPT, and Text-Davinci-003 respectively.

\noindent \textbf{Baselines} \enspace
The following models, including small-scale PLMs and LLMs, have been compared in our experiments.

For small-scale PLMs, we first select \textit{BERT} \cite{devlin-etal-2019-bert} and \textit{RoBERTa} \cite{liu2019roberta}. 
In addition, we also experiment with three strong and relevant models: 
(1) \textit{Hierarchical}~\cite{Adhikari2019DocBERTBF} which uses a pretrained model to encode different chunks of the document, and sets an additional BILSTM model to aggregate representations;
(2) \textit{SIEF}~\cite{xu-etal-2022-document} which proposes to randomly remove the useless sentences for prediction;
(3) \textit{SCS-EERE}~\cite{Man_Ngo_Van_Nguyen_2022} which selects a sentence set for each event pair for prediction.

The evaluated LLMs include: (1) \textit{GPT-4} which is an advanced and improved iteration of the GPT series, demonstrating human-level performance and significant enhancements in various aspects; 
(2) \textit{ChatGPT} which is an advanced conversational AI model, is able to provide contextually relevant and coherent responses aligned with human expectation; 
(3) \textit{Text-Davinci-003} which is a variant of GPT-3.5 series, offering
improved performance over GPT-3 through further instruction tuning.

\begin{table}[t]
    \centering
    \begin{tabular}{l p{0.9cm}<{\centering}p{0.9cm}<{\centering}p{0.9cm}<{\centering}}
         \toprule
        Methods&P&R&F1\\
         \midrule
         \multicolumn{4}{c}{MAVEN-ERE}\\
         \midrule
         BERT&$27.5$&$24.7$&$26.8$\\
         RoBERTa&$29.8$&$25.6$&$27.5$\\
         Hierarchical&$28.4$&$25.4$&$27.1$\\
         SIEF&$30.2$&$26.4$&$28.7$\\
         \midrule
         TacoERE (PLMs)&$\textbf{31.8}$&$\textbf{28.9}$&$\textbf{30.6}$\\
         \midrule
        \multicolumn{4}{c}{HiEve}\\
         \midrule
         BERT&$19.8$&$15.2$&$16.3$\\
         RoBERTa&$20.2$&$16.1$&$17.8$\\
         Hierarchical&$21.4$&$17.3$&$16.7$\\
         SIEF&$21.8$&$17.4$&$18.6$\\
         SCS-EERE&$20.6$&$\textbf{19.7}$&$19.2$\\
         \midrule
         TacoERE (PLMs)&$\textbf{22.6}$&$19.5$&$\textbf{20.8}$\\
         \bottomrule         
    \end{tabular}
    \caption{Model performance of subevent relation on MAVEN-ERE and HiEve.}
    \label{tab:subevent_per}
\end{table}

\subsection{Overall Results}
\label{exp_overallresults}

We report the results of causal relation on MAVEN-ERE and EventStoryLine, respectively. Subevent relation performance is evaluated on MAVEN-ERE and HiEve, respectively.

\textbf{Performance of causal relation} \enspace
The results are shown in Table \ref{tab:causal_per}, TacoERE (PLMs) means the implementation of our compression-then-extraction method with PLMs. 
Experimental results demonstrate that TacoERE (PLMs) outperforms all the baselines on both MAVEN-ERE and EventStoryLine datasets. We also have the following four observations: 
(1) compared with pretrained model BERT, TacoERE (PLMs) achieves 4.2\% improvements of F1-score on MAVEN-ERE, and 3.6\% on EventStoryLine data; 
(2) compared with Hierarchical, which models different chunks of document, TacoERE (PLMs) achieves improved performance. This validates the effectiveness of cluster summarization, which can further mitigate information redundancy and event distance; 
(3) compared with SIEF, we note that TacoERE (PLMs) achieves better performance, even though SIEF is designed for modeling the important content by removing sentences from the original document. This indicates that our method equipped with compression-then-extraction can effectively alleviate the problem of long-range dependencies; %, and more detailed analysis is included in following section; 
(4) SCS-EERE performs better than Hierarchical and SIEF. It adopts a straightforward idea that selects a set of sentences for each event pair. However, this might cost too much computation time when applied to large-scale datasets, and we directly predict all  event relations in a document at one time.

\textbf{Performance of subevent relation} \enspace
Table \ref{tab:subevent_per} presents the detailed results on MAVEN-ERE and HiEve, and our method again achieves better performance than all the baselines.
Our method improves upon the pretrained model BERT by 3.8\% and 4.5\% in terms of F1-score on both datasets, respectively.
In all, such performance on Table \ref{tab:causal_per} and \ref{tab:subevent_per} clearly demonstrates the benefits of compression-then-extraction on event relation extraction. 

% In following parts, we further analyze the results of a variety of experiments on MAVEN-ERE to validate TacoERE.

\begin{figure}[t]
  \centering
  \includegraphics[width=0.7\linewidth]{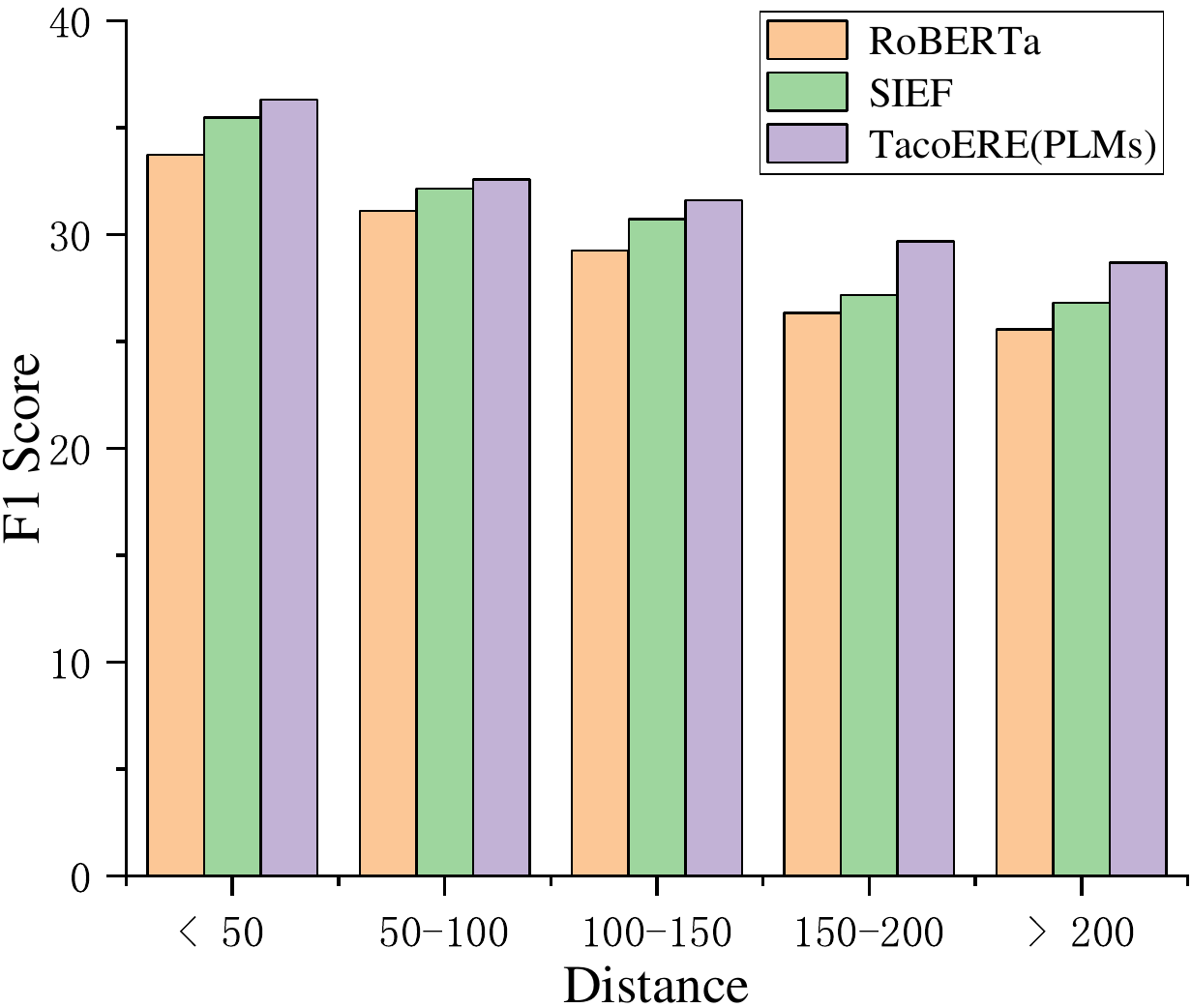}\\
  \caption{Model performance on different distance between related events (measured in \#words). }
  \label{fig:long_distance}
\end{figure}

\subsection{Impact on Event Distance}

To better understand the contributions of our method on long-range dependencies, we show the performance under different word distances between related events in Figure \ref{fig:long_distance}.

Compared with existing methods, our method achieves a certain improvement in dealing with long-range dependencies.
%From the Figure, 
We can also find:
(1) as the word distance continuously increases, the improvement of our method over the baselines shows an overall monotonic upward trend; 
(2) the overall F1-score on event pairs with long distance is much lower than that of short ones, indicating the challenge of long-range dependency; 
(3) particularly, compared with the two baselines which directly models the entire document (RoBERTa), or compresses the document by removing irrelevant sentences (SIEF), our method obtain a larger improvement, especially when the word distance of related events is greater than or equal to 4, which can further indicate that our TacoERE (PLMs) can better handle the long-range dependencies.

\begin{table}[t]
    \centering
    \begin{tabular}{l|cccc}
         \toprule
          Methods&EP&P&R&F1\\
         \midrule
         OneSum&$27.3$&$32.3$&$30.4$&$31.8$\\
         AvgSum&$50.4$&$33.7$&$31.4$&$32.5$\\
         TacoERE (PLMs)&$78.1$&$34.8$&$32.4$&$34.1$\\  
         \bottomrule
    \end{tabular}
    \caption{Model performance on different document compression strategies. EP is the ratio of events which has relations in summaries to document.}
    \label{tab:documentcompression_strategy}
\end{table}

\subsection{Document Compression Evaluation}
\label{exp_documentcompression}

In this section, we try different strategies to obtain the cluster and evaluate the performance to verify the effectiveness of our document compression.
We consider the following two strategies:
(1) ``\textit{OneSum}'' means only generating one summary for each document;
(2) ``\textit{AvgSum}'' refers to splitting document evenly into chunks based on sentence, and then generating summary for each chunk.

The results are shown in Table \ref{tab:documentcompression_strategy}, and we have the following three observations:
(1) compared with the other two methods, i.e., \textit{OneSum} and \textit{AvgSum}, our proposed TacoERE (PLMs) achieves the best performance;
(2) \textit{AvgSum} obtains second-best results.
However, the chunks are independent to each other in their setting, which means they ignore the relation information among chunks. This may prevent it from achieving better performance;
(3) for the metric EP, \textit{OneSum} only preserves 27.3\% events of the document, which is due to the fact that summary mainly focuses on the important content of the document and the length is relatively short compared to the input document. As a whole, model performance gradually gets better as more events are preserved.

\subsection{Ablation Study}
\label{exp_ablationstudy}

In addition to the document compression strategy, we also conduct experiments to ablate intra-clusters (\textit{w/o intra-clusters}), inter-clusters (\textit{w/o inter-clusters}) and cluster summarization (\textit{w/o summarization}) to understand their contributions. 
The results are shown in Table \ref{tab:ablation_study}. We can observe that: 
(1) compared the results of \textit{w/o (without) intra-clusters} with \textit{w/o inter-clusters}, we can find that the overall model works better. The use of intra- and inter-clusters helps the model to better understand the event relations both within and among the sentences;
(2) the performance of TacoERE drops on the three variations, which proves that both variations have contributed to the overall performance;
(3) removing \textit{w/o intra-clusters} causes a sharp performance drop compared to \textit{w/o inter-clusters}.

\begin{table}[t]
    \centering
    \begin{tabular}{l|p{0.9cm}<{\centering} p{0.9cm}<{\centering} p{0.9cm}<{\centering}}
         \toprule
         \multirow{2}*{Methods}&\multicolumn{3}{c}{MAVEN-ERE}\\
         \cmidrule{2-4}
          &P&R&F1\\
         \midrule
         TacoERE (PLMs)&$34.8$&$32.4$&$34.1$\\
         
         w/o intra-clusters&$32.4$&$30.9$&$32.3$\\
         
         w/o inter-clusters&$32.7$&$31.2$&$32.8$\\
         w/o summarization&$31.8$&$31.3$&$31.9$\\
         \bottomrule
    \end{tabular}
    \caption{Ablation study.}
    \label{tab:ablation_study}
\end{table}

\begin{table*}[t]
    \centering
    \begin{tabular}{l|p{0.9cm}<{\centering}p{0.9cm}<{\centering}p{0.9cm}<{\centering}|p{0.9cm}<{\centering}p{0.9cm}<{\centering}p{0.9cm}<{\centering}|p{0.9cm}<{\centering}p{0.9cm}<{\centering}p{0.9cm}<{\centering}}
         \hline
         \multirow{2}*{Methods}&\multicolumn{3}{c|}{Text-Davinci-003}&\multicolumn{3}{c|}{ChatGPT}&\multicolumn{3}{c}{GPT-4}\\
         %\cline{2-10}
          &P&R&F1&P&R&F1&P&R&F1\\
         \midrule
         Document &$13.8$&$6.2$&$8.5$&$21.7$&$32.2$&$25.9$&$27.1$&$41.5$&$32.8$\\
         Sentence Pair&$21.9$&$7.1$&$10.7$&$24.3$&$31.2$&$27.3$&$33.4$&$38.6$&$35.7$\\
         \midrule
         Document Clustering&$17.3$&$8.1$&$10.9$&$24.6$&$32.9$&$28.2$&$31.9$&$47.1$&$38.1$\\         
         TacoERE (LLMs)&$\textbf{30.2}$&$\textbf{8.9}$&$\textbf{13.8}$&$\textbf{31.3}$&$\textbf{45.6}$&$\textbf{37.1}$&$\textbf{38.9}$&$\textbf{45.5}$&$\textbf{41.9}$\\
         \hline
    \end{tabular}
    \caption{Model performance of causal relation on different LLMs. Experiments are under 2-shot setting.}
    \label{tab:llmresults}
\end{table*}

\begin{figure*}[h]
  \centering
  \includegraphics[width=\linewidth]{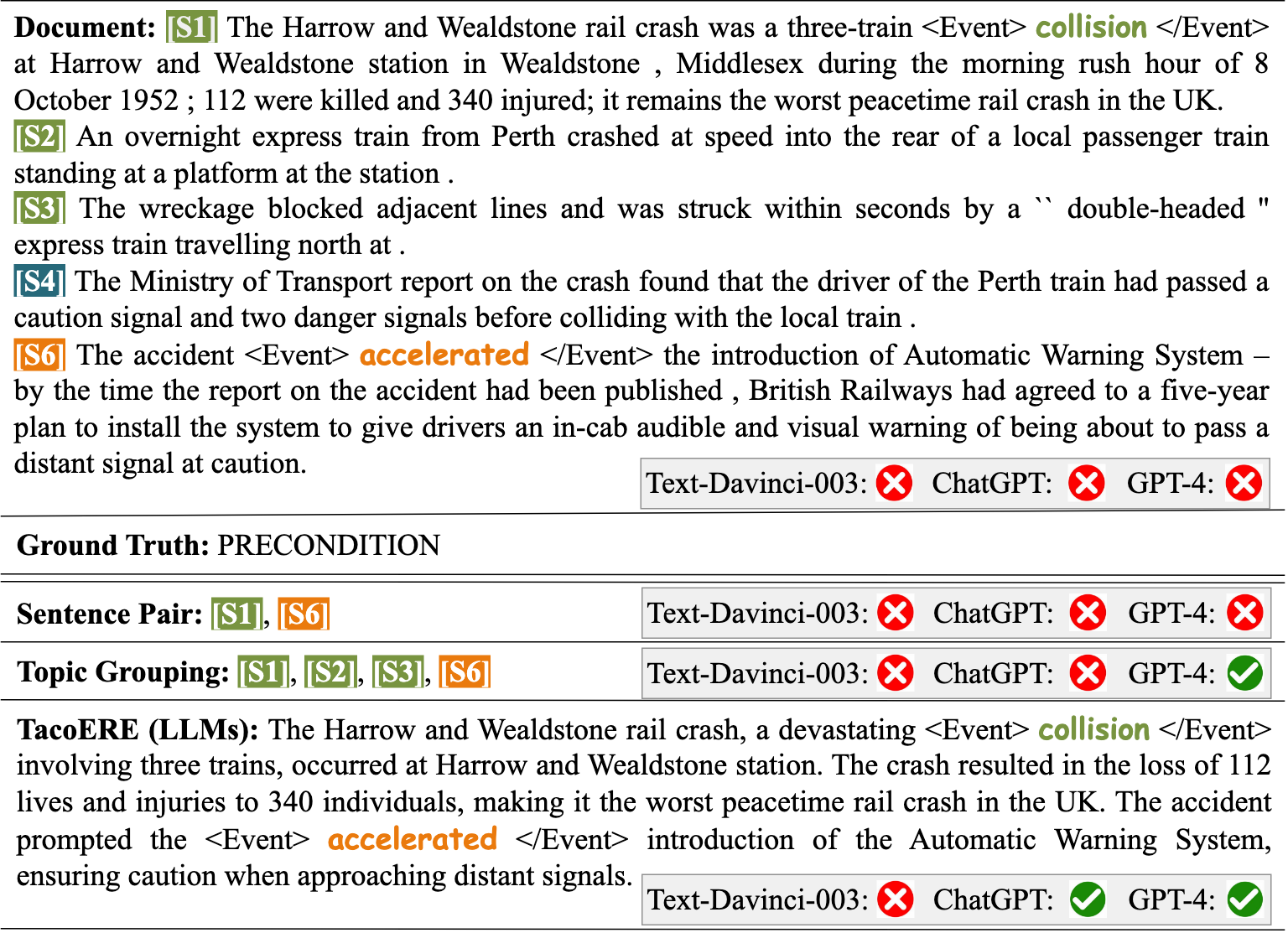}\\
  \caption{Case analysis of relation prediction.}
  \label{tab:case2}
\end{figure*}

\subsection{Evaluation on LLMs}
\label{exp_llms}

The aforementioned series of experiments has highlighted the efficacy of our TacoERE, based on PLMs, in enhancing ERE performance. To further validate its effectiveness and robustness, we conduct extensive experiments on LLMs. Illustrated in Figure \ref{meth_struc}, our framework primarily comprises three key components: Document Clustering, Cluster Summarization, and Relation Prediction. We introduce TacoERE (LLMs), which directly leverages LLMs to implement these three components. To thoroughly validate our approach, we configure the task to enable relation prediction between event pairs individually. For testing, we randomly sample 50 documents from MAVEN-ERE, resulting in 646 causal relations. We compare TacoERE (LLMs) with three variants: (1) \textit{Document}, which involves utilizing the entire document to predict relations; (2) \textit{Sentence Pair}, which entails using sentences containing the event pairs to predict relations; (3) \textit{Document Clustering}, which involves using sentences within a cluster to predict relations.

The results are shown in Table \ref{tab:llmresults}, we notice that our TacoERE (LLMs) achieves the best performance across all three models, with improvements of 11.2\% and 9.1\% on ChatGPT and GPT-4, respectively. We also have the following four observations: 
(1) from model perspective, GTP-4 achieves the highest F1 score of 41.9\%, followed by ChatGPT;
(2) compared with \textit{Document}, \textit{Sentence Pair} obtains better results, indicating that predicting relation between events does not depend on the whole document; 
(3) compared with \textit{Document} and \textit{Sentence Pair}, \textit{Document Clustering} achieves improved results, indicating that our method can reduce redundant information while retaining useful information for relation prediction; 
(4) compared with \textit{Document Clustering}, our TacoERE (LLMs) achieves the best performance, suggesting that reducing redundant information and shortening event distance can further facilitate the improvement of performance.

\subsection{Case Study}

We display a case in Figure \ref{tab:case2} to quantitatively analyze the model prediction by our TacoERE (LLMs) and different comparison modules, such as \textit{Document}, \textit{Sentence Pair}, and \textit{Document Clustering}. TacoERE (LLMs) means we use the content from \textit{Cluster Summarization} to predict relations. 
We can see, for each event pair, the prediction does not rely on the whole document. Some dependency information may be ignored by using \textit{Document Clustering} for prediction. Our proposed TacoERE (LLMs) with compression-then-extraction is an effective means to enhance event relation extraction.

\section{Related Work}

\subsection{Event Relation Extraction}
Event Relation Extraction is a challenging task in natural language processing, especially for events that are scattered in different sentences \cite{gao-etal-2019-modeling,chen-etal-2022-ergo}. 
Recently, deep learning based methods are becoming the mainstream \cite{cao-etal-2021-knowledge,xu-etal-2021-discriminative}, 
and extensive explorations have been made, such as joint reasoning methods which extract multiple relations simultaneously \cite{ning-etal-2018-joint,han-etal-2019-joint}, and graph-based methods which use event mentions as nodes and model document as graph~\cite{tran-phu-nguyen-2021-graph,fan_eci_2022,guo-etal-2023-eventoa}. However, these works use document as input, which cannot well handle the long-range dependency problem. In contrast, we improve event relation extraction by processing document in advance with cluster-aware compression. 

Currently, a series of LLMs have been developed, such as GPT series, LaMDA~\cite{thoppilan2022lamda}, and PaLM~\cite{chowdhery2022palm}, and have achieved remarkable performance in various fields. Among them, GPT series, i.e., GPT-4 and ChatGPT, is undoubtedly the most popular work. Thus, to verify the effectiveness of our method, we conduct extensive experiments on them.

\subsection{Controlled Text Summarization}
With the development of deep learning and the increasing demand for generation quality, increasing studies are focusing on controlled text summarization~\cite{dou-etal-2021-gsum,guan-etal-2021-frame}, such as designing copy mechanism to directly copy important word from input~\cite{see-etal-2017-get}, extracting actual fact triples for modeling~\cite{faithful_cao_2017}, extracting templates from the training data to guide the summarization generation~\cite{wang-etal-2019-biset}.
However, these methods typically design an additional module or using existing third-party content selectors.
On the contrary, our method adopts the events in document as intermediate representation to better guide the summary generation.

\section{Conclusion}

In this paper, we propose a novel cluster-aware compression method for event relation extraction, namely, TacoERE, which explores a compression-then-extraction paradigm to extract event relations. 
TacoERE first splits document into intra- and inter-clusters to allow the modeling of dependencies without considering event distance among sentences. Then, cluster summarization is adopted to simplify and highlight the important text of clusters for further mitigating information redundancy and event distance. Extensive experiments have been conducted on both small-scale PLMs such as RoBERTa, and LLMs such as ChatGPT and GPT-4. Experimental results demonstrate that our proposed TacoERE with compression-then-extraction is an effective method for augmenting event relation extraction.

\section{Bibliographical References}

\bibliographystyle{lrec-coling2024-natbib}
\bibliography{custom}

\section{Language Resource References}

\bibliographystylelanguageresource{lrec-coling2024-natbib}
\bibliographylanguageresource{languageresource}

\end{document}